\documentclass[journal]{IEEEtran}
\usepackage[pdftex]{graphicx}
\usepackage{cite}
\usepackage{amsmath,amssymb,amsfonts}
\usepackage{fixltx2e}
\usepackage{array}
\usepackage{algorithmic}
\usepackage{float}
\usepackage{xcolor}
\usepackage{orcidlink}
\usepackage{stfloats}
\usepackage{url}

\begin{document}
\title{Deep Learning Assisted Inertial Dead Reckoning and Fusion}

\author{Dror~Hurwitz \orcidlink{0000-0003-4183-1225}, Nadav~Cohen \orcidlink{0000-0002-8249-0239} 
        and~Itzik~Klein \orcidlink{0000-0001-7846-0654}
\thanks{D. Hurwitz, N. Cohen and I. Klein are with the Hatter Department of Marine Technologies, Charney School of Marine Sciences, University of Haifa, Israel.\\ Email: drorhurw@gmail.com (D.Hurwitz), ncohe140@campus.haifa.ac.il (N.Cohen), kitzik@univ.haifa.ac.il (I.Klein)}
}
	
		
	
	%
	%

\markboth{Preprint}%
{Shell \MakeLowercase{\textit{et al.}}: EKF-QuadNet: a new sensor fusing way for drones}

\maketitle

\begin{abstract}
The interest in mobile platforms across a variety of applications has increased significantly in recent years. One of the reasons is the ability to achieve accurate navigation by using low-cost sensors. To this end, inertial sensors are fused with global navigation satellite systems (GNSS) signals. GNSS outages during platform operation can result in pure inertial navigation, causing the navigation solution to drift. 
In such situations, periodic trajectories with dedicated algorithms were suggested to mitigate the drift. With periodic dynamics, inertial deep learning approaches can capture the motion more accurately and provide accurate dead-reckoning for drones and mobile robots. In this paper, we propose approaches to extend deep learning-assisted inertial sensing and fusion capabilities during periodic motion.   We begin by demonstrating that fusion between GNSS and inertial sensors in periodic trajectories achieves better accuracy compared to straight-line trajectories. Next, we propose an empowered network architecture to accurately regress the change in distance of the platform. Utilizing this network, we drive a hybrid approach for a neural-inertial fusion filter. Finally, we utilize this approach for situations when GNSS is available and show its benefits. A dataset of 337 minutes of data collected from inertial sensors mounted on a mobile robot and a quadrotor is used to evaluate our approaches.
\end{abstract}


%
\IEEEpeerreviewmaketitle
\section{Introduction}
\IEEEPARstart{T}{he} use of autonomous platforms in different operating environments has significantly increased in the past decade. Among these platforms are quadrotors, underwater and surface vehicles, and mobile robots. Those autonomous platforms are used in many applications such as agriculture, construction, healthcare, delivery, transportation, surveillance, and more \cite{idrissi2022review,sonugur2023review,dobrev2017steady,gao2018review,park2016bim,chen2017real}. \\
\noindent
An enabler for successful autonomous missions is the ability to achieve accurate navigation. Commonly, cameras or global navigation satellite systems (GNSS) receivers combined with inertial sensors are used to determine the navigation solution \cite{chang2015huber}. Due to noise and errors in the inertial sensors, the navigation solution drifts over time in pure inertial navigation scenarios \cite{titterton2004strapdown}. To circumvent the inertial drift, they are fused with GNSS or cameras in a nonlinear filter framework \cite{zhang2018intelligent, boguspayev2023comprehensive}.\\
Inertial navigation system (INS) fusion with GNSS has been studied for more than twenty years \cite{titterton2004strapdown, Farell2015} including topics of the fusion type {\cite{wang2019constrained}, \cite{chen2020estimate}}, observability {\cite{hong2008observability} \cite{li2023accuracy}}, multi-antenna GNSS~\cite{cai2019multi}\cite{yoder2023low}, lever-arm influence~\cite{s18072228}, and more. 
Several recent works have explored integrating neural networks and Kalman filters for position estimation with INS/GNSS in different scenarios.
\cite{du2023neural} proposed a system for land vehicles used in agriculture tasks. Their approach combines a temporal convolutional network (TCN) with a transformer architecture to estimate velocity using inertial data, while incorporating position updates from GNSS. For land vehicles, \cite{hosseinyalamdary2018deep} introduced a deep Kalman filter utilizing recurrent neural networks and long short term memoery networks. Their approach incorporates a modeling step that corrects the state vector.
\cite{wu2020predicting} proposed a method that uses a TCN to regress the parameters of measurement noise covariances and process noise covariances.
For quadrotor applications, \cite{or2022hybrid} developed an adaptive noise covariance approach based on data-driven learning with a 1-dimensional CNN.
\\
\noindent
However, in typical scenarios, the GNSS signals may be blocked, and the camera may suffer from poor lighting conditions. As a consequence, no external update is available, leading to pure inertial navigation and, consequently, rapid growth in positioning errors. To handle such behavior, the quadrotor dead reckoning (QDR) approach was recently proposed \cite{shurin2020qdr}.
The QDR method steams from pedestrian dead reckoning (PDR) approaches where wearable inertial sensors are employed to determine the position of the pedestrian \cite{hou2020pedestrian}. While in PDR, the periodic motion is a natural result of human movement, in QDR, periodic trajectories (PTS) are enforced to enable accurate navigation.  In this manner, the peak-to-peak (period) distance of the quadrotor can be accurately estimated, similar to how step-length is detected and estimated in PDR. Motivated by recent applications of deep-learning in navigation applications \cite{cohen2023inertial, klein2022data}, the QuadNet framework was suggested to replace the model-based QDR \cite{shurin2022quadnet}. QuadNet is a hybrid deep-learning framework employing only inertial sensor measurements to estimate the three-dimensional position vector. Using regression neural networks, QuadNet calculates the quadrotor's change in distance and altitude in any desired time window, while the heading angle is obtained from the inertial navigation system mechanism. Later, QuadNet was generalized to allow the usage of multiple inertial measurement units (IMUs) to increase the robustness and improve the positioning accuracy \cite{hurwitz2023quadrotor}. As the quadrotor maneuvers during PTS, battery consumption is expected to be higher than a straight-line trajectory. In a recent study for quadrotors operating indoors, it was shown that for an equal length trajectory, PTS requires additional battery consumption of approximately $15\%$ \cite{aizelman2024quadrotor}.   \\
\noindent
Motivated by QDR, PTS was also employed in the development of a mobile robot pure inertial framework (MoRPI) \cite{etzion2023morpi}. There, closed-form analytical solutions are derived to show that MoRPI produces lower position error compared to the classical pure inertial solution. Later, PTS was adopted also for glider navigation \cite{weizman2023enhancement}. There, GliderNet, an end-to-end framework utilizing only accelerometer readings, was defined to regress the glider depth and distance. GliderNet was evaluated only on stimulative data. \\
\noindent
In this paper, we propose approaches to extend deep learning assisted inertial sensing and INS/GNSS fusion capabilities during periodic motion. The contributions of this paper are as follows:  
\begin{enumerate}
	\item \textbf{PTS Trajectories for INS/GNSS Fusion}: We show that fusion between GNSS and inertial sensors using PTS in an EKF framework achieves better accuracy, compared to straight line trajectories. 
	\item \textbf{Mini-QuadNet}:  A deep-learning network, based on QuadNet, for the change in distance regression task, is proposed. With proper design, we shrink the number of layers and parameters in the original QuadNet network and present the Mini-QuadNet. In addition to the reduced complexity,  Mini-QuadNet shows superior performance over QuadNet.
	\item \textbf{Mini-QuadNet/INS/GNSS Fusion}:  A hybrid approach for neural-inertial fusion in the EKF framework is derived. There, both GNSS and Mini-QuadNet measurements are employed to update the INS. We highlight the addition of Mini-QuadNet measurements in empowering GNSS/INS fusion during PTS motion.
	\item \textbf{Mini-QuadNet/INS Fusion}:  We demonstrate that Mini-QuadNet measurements can aid the INS in situations of GNSS outages. To that end, Mini-QuadNet measurements are used as external updates to the INS in an EKF framework.
\end{enumerate}

\noindent
We evaluate our approach on real-world data collected from a mobile robot and a quadrotor.  The data includes inertial readings recorded by four and five Movella DOT inertial measurement units (IMU) mounted on the quadrotor and mobile robot, respectively. In total 49 trajectories with a duration of 82.2 minutes were evaluated for each IMU. The ground-truth (GT) trajectories, recorded using a real time kinematic (RTK) GNSS receiver, are also provided in our dataset.
\\
\noindent
The rest of the paper is organized as follows: Section~\ref{sec:PF} gives the QuadNet framework and the EKF navigation filter. Section~\ref{sec:PA} presents our proposed approach, including Mini-Quadnet and its measurement update model in the EKF. Next, in Section~\ref{sec:AER}, a detailed description of the quadrotor and mobile robot datasets is provided, as well as the experimental results. Lastly, Section~\ref{sec:con} derives the conclusions of this research.
\section{Problem Formulation}\label{sec:PF}
\noindent
In this section, we describe the QuadNet method and present the error-state EKF, which is later implemented as part of our proposed approach.
\subsection{QuadNet Framework with PTS}
\begin{figure}[h!]
			  \begin{center}
	\includegraphics[width=\columnwidth ]{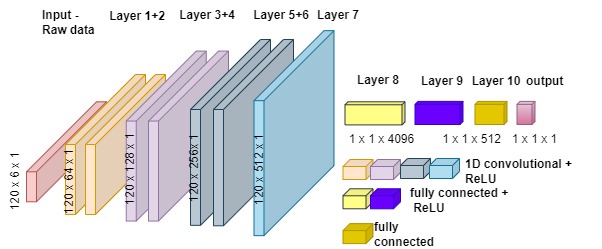}%
			\end{center}
	\caption{QuadNet architecture as presented in \cite{shurin2022quadnet}. It consists of convolutional and fully connected layers to output the change in the platform's distance.}
	\label{fig:QuadNetArchi}
\end{figure}

\noindent
QuadNet utilizes neural networks to estimate the quadrotor's position vector during motion in PTS. Inertial sensor readings are fed into a regression network to predict the change in distance and altitude. Simultaneously, an attitude and heading reference system (AHRS) algorithm is applied to determine the heading angle. Finally, the estimated change in distance and altitude, and the heading angle are introduced into the equations of motion to calculate the platform's position.\\
\noindent
The basic QuadNet architecture comprises of seven one-dimensional convolutional layers and three fully connected (FC) layers as can be seen in Fig. \ref{fig:QuadNetArchi}. The raw inertial (accelerometer and gyroscope) measurements are employed as the input to the network. The output of the network is the change in altitude or distance.  This architecture leverages the complementary strengths of convolutional and FC layers. CNNs excel at capturing spatial relationships within data due to their inherent ability to process information locally. This allows them enhanced abilities to cope with sensor noise and facilitates the construction of larger networks with fewer parameters, leading to easier training and reduced overfitting. On the other hand, FC layers excel at learning global features, complementing the capabilities of convolutional layers. \\
\noindent
The cross-correlation operation in a one-dimensional convolutional layer is defined by:
\begin{equation}
    (\mathbf{\chi} \star \mathbf{w})(t) := \int_{-\infty}^{\infty} \mathbf{\chi}(\tau) \mathbf{w}(\tau - t) \, d\tau
\end{equation}
where $\mathbf{\chi}$ is the input, $\mathbf{w}$ is the filter (or weights), and $\star$ is the cross-correlation operator. The filter $\mathbf{w}$ is referred to as the weight matrix, which is updated during the training process.
In the discrete case, which is used in neural networks, this can be written as:
\begin{equation}
    (\mathbf{\chi} \star \mathbf{w})(t) := \sum_{m=0}^{M-1} \mathbf{\chi}(t+m) \mathbf{w}(m)
\end{equation}
where the sum is taken over the length of the filter $M$. The cross-correlation operation slides the filter $\mathbf{w}$ over the input $\mathbf{\chi}$ and computes the element-wise product of the overlapping elements, summing them to produce the output at each position $t$.\\
Let $\textbf{x}$ denote the input layer, then the first hidden layer is defined by:
\begin{equation}\label{eq:layer}
	\textbf{h}^{(1)} = \textbf{g}^{(1)}(\textbf{w}^{(1)^T}\textbf{x} + \textbf{b}^{(1)})
\end{equation}
where $\textbf{b}$ is the bias vector, which is also updated during the training process, and $\textbf{g}$ is a nonlinear activation function. In following layers, the weights and biases create a mapping between the neurons in the current layer and the neurons from the previous layer while the activation function adds nonlinearities to enrich the network performance. QuadNet employs  the rectified linear unit (ReLU) \cite{hara2015analysis}:
\begin{equation}
	\textbf{g(z)} = max\{\mathbf{0},\textbf{z}\}
\end{equation}
where $\textbf{g(z)}$ is the input to the activation function. The i-th layer is defined similarly to \ref{eq:layer}, replacing the input by the output of the i-1 layer:
\begin{equation}
	\textbf{h}^{(i)} = \textbf{g}^{(i)}(\textbf{w}^{(i)^T}\textbf{h}^{(i-1)} + \textbf{b}^{(i)})
\end{equation}
\subsection{Navigation Filter}
\noindent
Due to the nonlinear nature of the INS equations of motion, their fusion with external sensors necessitates the use of nonlinear filters. Commonly, the error-state EKF is used for this purpose with a 15-state vector \cite{Farell2015, groves2013, noureldin2012fundamentals}:
\begin{equation}
	\delta\bold{x} = \begin{bmatrix}
		\delta \bold{p}^T& \delta \bold{v}^T & \boldsymbol{\epsilon}^T & \bold{b}_a^T & \bold{b}_g^T 
	\end{bmatrix}^T \in \mathbb{R}^{15}
	\label{eq:states}
\end{equation}
where $\delta \bold{x}$ is the error state vector, $\delta \bold{p}$ is the position error-state vector, $\delta \bold{v}$ is the velocity error state vector,  $ \bold{\epsilon}$ is the misalignment error-states, $\bold{b}_a$ is the accelerometer bias vector, and $\bold{b}_g$ is the gyroscope bias vector. It is assumed that the dominant error in the inertial sensors is the bias, and it is modeled as a random walk process. Thus, the linearized state-space error model is
\begin{equation}
	\label{ode}
	\delta\bold{\dot{x}} = \bold{F}\delta\bold{{x}} +\bold{G}\delta\bold{{w}}
\end{equation}
where $\bold{F}$ is the system matrix, $\bold{G}$ is the shaping matrix, and $\bold{\delta{w}}$ is the noise vector. The system matrix is defined as
\begin{equation}
	\label{systemMatrix}
	\bold{F} = \begin{bmatrix}
		\bold{F}_{pp}&\bold{F}_{pv}&\bold{0}_{3\times3}&\bold{0}_{3\times3}&\bold{0}_{3\times3}\\
		\bold{F}_{vp}&\bold{F}_{vv}&\bold{F}_{v\epsilon}&\bold{T}_b^n&\bold{0}_{3\times3}\\
		\bold{F}_{\epsilon p}&\bold{F}_{\epsilon v}&\bold{F}_{\epsilon \epsilon}&\bold{0}_{3\times3}&\bold{T}_b^n\\
		\bold{0}_{3\times3}&\bold{0}_{3\times3}&\bold{0}_{3\times3}&\bold{0}_{3\times3}&\bold{0}_{3\times3}\\
		\bold{0}_{3\times3}&\bold{0}_{3\times3}&\bold{0}_{3\times3}&\bold{0}_{3\times3}&\bold{0}_{3\times3}
	\end{bmatrix}
\end{equation}
where $\bold{T_b^n}$ is the transformation matrix between the body to the navigation frame. The other system submatrices $\bold{F_{ij}}\in\mathbb{R}^{3\times3}$ are derived analytically and can be found in detail in \cite{Farell2015, groves2013}.  \\
The shaping matrix associated with the model is
\begin{equation}
	\bold{G} = 
	\begin{bmatrix}
		\bold{0}_{3\times3} & \bold{0}_{3\times3} & \bold{0}_{3\times3} & \bold{0}_{3\times3} \\
		\bold{T_b^n} & \bold{0}_{3x3} & \bold{0}_{3x3} & \bold{0}_{3x3} \\
		\bold{0}_{3\times3} & \bold{T_b^n} & \bold{0}_{3\times3} & \bold{0}_{3\times3} \\
		\bold{0}_{3\times3} & \bold{0}_{3\times3} & \bold{I}_{3} & \bold{0}_{3\times3}\\
		\bold{0}_{3\times3} & \bold{0}_{3\times3} & \bold{0}_{3\times3} & \bold{I}_{3}
	\end{bmatrix}
\end{equation}
and $\bold{\delta{w}}$ is a zero-mean, white Gaussian noise vector consisting of the four different noise sources in the system
\begin{equation}
	\bold{\delta{w}} =
	\begin{bmatrix}
		\bold{w}_a & \bold{w}_g & \bold{w}_{ba} & \bold{w}_{bg}
	\end{bmatrix} ^T
\end{equation}	
where the zero-mean white Gaussian noise of the accelerometers and gyroscopes are denoted by $\bold{w}_a$ and $\bold{w}_g$, respectively. Similarly, $\bold{w}_{ba}$ and $\bold{w}_{bg}$ represent the random walk noises of the inertial sensors.\\
\noindent
The EKF closed-loop implementation algorithm prediction phase is~\cite{Farell2015}
\begin{equation}
	\label{EKFIC}
	\bold{\delta\hat{x}_{k}^-} = \bold{0}
\end{equation}
\begin{equation}
	\label{EKFstartLoop}
	\boldsymbol{\Phi}_k = \textbf{I}_{15} + \textbf{F}_k t_s
\end{equation}
\begin{equation}
	\label{EKFCovariance}
	\textbf{P}_{k}^- = \boldsymbol{\Phi}_{k-1}\textbf{P}_{k-1}^+\boldsymbol{\Phi}^T_{k-1} + \textbf{Q}_{k-1}
\end{equation}
where $\bold{\delta\hat{x}_{k}^-}$ is the \emph{a priori} error-state, $\boldsymbol{\Phi}_k$, is the state transition matrix approximated to its first order using the system matrix $\textbf{F}_k$, defined in \eqref{systemMatrix}, and $t_s$ is the step-size. The error state covariance matrix is denoted by $\textbf{P}^{-}_k$ and $\textbf{Q}_k$ is the process noise covariance matrix. \\
The EKF update phase is:
\begin{equation}
	\label{eq:measure}
	\bold{\delta\tilde{ y}_k} = \bold{z_k} - \textbf{H}_k \bold{\hat{{x}}_{k}^-}
\end{equation}
\begin{equation}
	\label{eq:KalmanGain}
	\textbf{K}_k = \textbf{P}_{k}^- \textbf{H}_k^T (\textbf{H}_{k}\textbf{P}_{k}^-\textbf{H}_k^T +\textbf{R}_k)^{-1}
\end{equation}
\begin{equation}
	\label{eq:UpdateState}
	\bold{\delta\hat{x}_{k}^+} = \textbf{K}_k\delta\bold{\tilde{y}_k}
\end{equation}
\begin{equation}
	\label{eq:CovarinaceUpd}
	\textbf{P}_{k}^+ = [\textbf{I} - \textbf{K}_k\textbf{H}_k]\textbf{P}_{k}^-
\end{equation}
where  $\bold{\delta\hat{x}_{k}^+}$ is the \emph{posteriori} error-state vector, $\bold{{z}_k}$ is the measurement vector, $\bold{\delta\tilde{ y}_k}$ is the measurement residual reflecting the difference between the predicted measurement and the actual measurement, $\textbf{K}_k$ is the Kalman gain matrix which determines the weight given to the new measurement during the update step, balancing between the prediction and the new information. The measurement noise covariance matrix is denoted as $\textbf{R}_k$, $\textbf{H}_k$, is the measurement matrix, and $\textbf{P}_{k}^+$ is the \emph{posteriori} error-state covariance matrix.\\
When  GNSS position vector measurement is used to update the filter, the measurement matrix is defined by
\begin{equation}
	\textbf{H}_K^{GNSS}=
	\begin{bmatrix}
		\textbf{I}_{3} & \textbf{0} _{3\times 3} & \textbf{0} _{3\times 3} & \textbf{0} _{3\times 3} & \textbf{0} _{3\times 3}
	\end{bmatrix}
	\label{eq:Hgnss}
\end{equation}
\section{Proposed Approach}\label{sec:PA}
\noindent
In order to enable neural networks to be adopted in low performance processors, we propose Mini-QuadNet, which provides a reduced size of network while maintaining and even improving the baseline QuadNet performance. Next, we describe our approach for neural-inertial fusion with Mini-QuadNet updates in an error-state implementation of the extended Kalman filter. 
\subsection{Mini-QuadNet}
\noindent
Here, we introduce Mini QuadNet (MQN), a neural network based on the baseline QuadNet architecture. After some analysis, the number of CNN layers was reduced from seven to five by eliminating the two layers with an input of 120x128x1 (Layer 3+4 in the baseline architecture) and an output of 120x256x1 (Layer 5 in the baseline architecture). Additionally, two FC layers, the first with and input and output size of 4096 (Layer 8 in the baseline architecture) connected to another layer with an output of 512 (Layer 9 in the baseline architecture) were removed. As a consequence, the number of parameters was reduced from 272 million parameters to 32 million parameters, reflecting a $88\%$ parameters reduction. As a consequence, the weights size was reduced from 1 GB to only 126 MB. Following the network size reduction, the inference time was also reduced by a factor of five. When tested on an i5-1235U 1.30 GHz CPU, the inference time was reduced from 0.1 to 0.02 seconds.
\begin{figure}[h!]
		  \begin{center}
	\includegraphics[width=\columnwidth ]{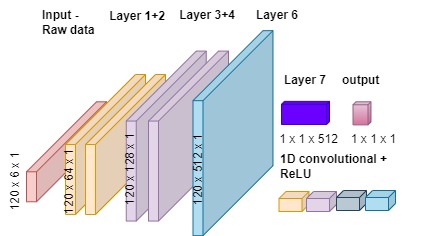}%
	  \end{center}
	\caption{The MQN network architecture.  The inertial readings are passed into a series of 1D-CNN and fully connected layers to output the change in distance.}
 \label{fig:ttttt}
\end{figure}
The hyperparameters used in the MQN network are an ADAM optimizer with a scheduler reduce on plateau with a decay of 0.7 and patience of 4. The number of epochs is 70, the batch size is 64, and the window size was set to contain 120 samples. At training, the stride was half the size of window while in testing the entire window (no overlapping) was used.\\
Two of the hyperparameters, the dropout at the last fully connected layer and learning rate, were optimized using Optuna \cite{optuna_2019}, with a Monte-Carlo of 50 trials. 
During the training process, the mean squared error (MSE) loss was utilized: 
\begin{equation}
		L(d_i, \hat{d}_i) = \frac{1}{N} \Sigma^{N}_{i=1} (d_i - \hat{d}_i)^2
\end{equation}
where $d_i$ is the regressed distance value, $\hat{d}_i$ is the GT value and $N$ is the number windows in the training dataset.\\
Using the same MQN network and the above procedure, the platform's change in altitude during the window size duration can also be regressed. The only difference is the definition of the loss function, which now reads:
\begin{equation}
	L(\Delta h_i, \Delta \hat{h}_i) = \frac{1}{N} \Sigma^{N}_{i=1} (\Delta h_i - \Delta \hat{h}_i)^2
\end{equation}
where $\Delta h_i$ is the regressed change in altitude value, $\hat{h}_i$ is the GT value and $N$ is the number windows in the training dataset.\\
Given the regressed distance and change in altitude from MQN, the AHRS yaw angle, and the position initial conditions, the platform position components are propagated by:
\begin{eqnarray}\label{eq:pos_x}
		x_{k+1}  & =  & x_k + \hat{d}_k \cdot \cos(\psi_k) \\
\label{eq:pos_y}
		y_{k+1} & = & y_k  + \hat{d}_k \cdot \sin(\psi_k)\\
\label{eq:pos_z}
		z_{k+1} & =  & z_k + \Delta \hat{h}_k
\end{eqnarray}
where $\psi$ is the yaw angle and $x$, $y$ and $z$ are the  position vector components.
Equations \eqref{eq:pos_x}-\eqref{eq:pos_y} are valid for mobile robots and quadrotor inertial navigation. For the latter, the altitude information is provided by \eqref{eq:pos_z}.  
\subsection{Neural-Inertial Fusion}
\begin{figure}[h!]
	\begin{center}
		\includegraphics[width=\columnwidth ]{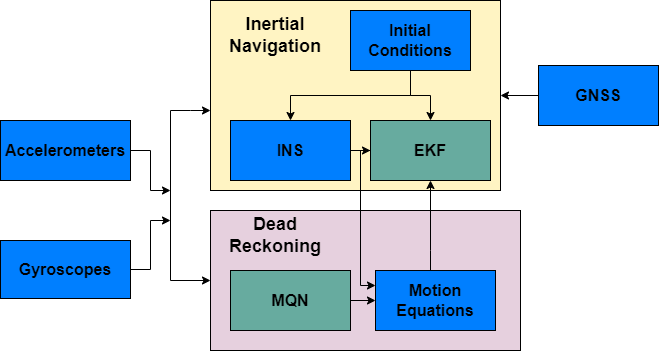}%
	\end{center}
	\caption{The neural inertial fusion block diagram. MQN-driven position, as well as GNSS updates, when available, are used as external updates to the navigation filter.}
	\label{blockDiagram}
\end{figure}

\noindent
We propose a neural inertial fusion by employing the MQN regressed position vector as an external update to the navigation filter. We leverage this type of fusion for situations when GNSS updates are unavailable and also as an add-on update when they are. A block diagram presenting the proposed approach is given in Fig. \ref{blockDiagram}. 
\noindent
The inertial sensors, namely the accelerometer and gyroscope, readings are fed into the dead reckoning part and the inertial navigation block. In the dead-reckoning block, those readings are fed into the MQN network to regress the change in distance and altitude of the platform during a predefined window size. Those estimates are plugged into the motion equations \eqref{eq:pos_x}-\eqref{eq:pos_z} to calculate the platform position vector. In parallel, the inertial readings are employed in the inertial navigation block to propagate the inertial navigation system equations and the navigation filter \eqref{EKFIC}-\eqref{eq:CovarinaceUpd}. If GNSS position measurements are available, they are inserted into the filter using the corresponding measurement matrix \eqref{eq:Hgnss}. \\
\noindent
Regardless of the availability of the GNSS measurements, the MQN-driven position vector is used to update the filer. To that end, we construct the position measurement as
\begin{equation}\label{eq:pos_mqn_fil}
	\bold{z^{\text{MQN}}_k} = \begin{bmatrix}
		x_{k} & y_{k} & z_{k}
	\end{bmatrix}^{T}
\end{equation}
with its corresponding measurement matrix
\begin{equation}
	\textbf{H}_k^{\text{MQN}}=
	\begin{bmatrix}
		\textbf{I}_{3} & \textbf{0} _{3\times 3} & \textbf{0} _{3\times 3} & \textbf{0} _{3\times 3} & \textbf{0} _{3\times 3}
	\end{bmatrix}
	\label{eq:HQuadNet}
\end{equation}
Finally, \eqref{eq:pos_mqn_fil}-\eqref{eq:HQuadNet} are introduced into the navigation filter update stage \eqref{eq:measure}-\eqref{eq:CovarinaceUpd}. \\
As explained above, the MQN position updates (measurement model) are based on the inertial readings as well as the INS equations of motion, which serve as the system model. Thus, there exists a correlation between the process (system) noise and measurement noise which now should be taken into account in the formulation of the navigation filter as shown in~\cite{simon2006optimal}. In this work, we assume that this process-measurement correlation matrix can be neglected. 
Also, to maximize the performance of our proposed approach it should be applied only on PTS trajectories. The nature of those trajectories enables the MQN network to accurately regress the platform position vector as we demonstrate in the next section.  
\section{Analysis and Experiment Results}\label{sec:AER}
\noindent
In this section, we evaluate our proposed approaches on two different platforms, a mobile robot and a quadrotor, using real-world recorded data. We first describe the datasets followed by analysis of the results.
\subsection{Inertial Datasets}
\begin{figure}[h!]
	\begin{center}
	\includegraphics[width=\columnwidth]{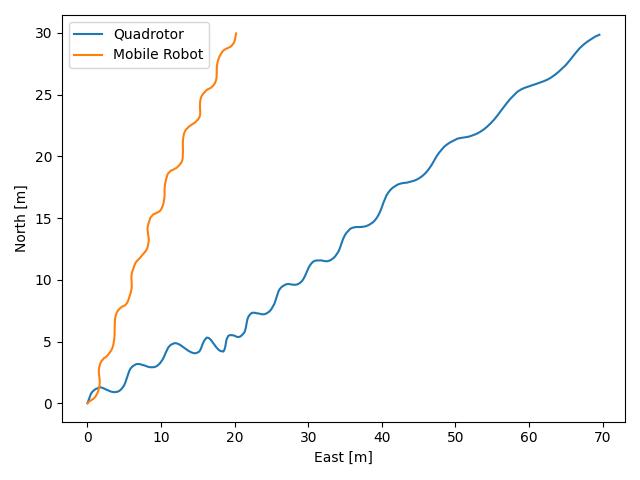}%
	\end{center}
	\caption{Examples of PTS trajectories, from our dataset, of a quadrotor (blue) and a mobile robot (orange).}
	\label{Fig:Phantom44}
\end{figure}

\noindent
Two datasets are employed in our analysis. The first was generated using a quadrotor, while the second was done by a remote-controlled car, both following PTS. In total, the datasets contain 337 minutes of inertial recordings and corresponding GT. Examples of PTS trajectories from our dataset of a quadrotor and a mobile robot are shown in Fig.~\ref{Fig:Phantom44}. It can be seen from the figure that the periodic motion is not symmetrical and that its period and amplitude change over the course of the trajectory, allowing us to evaluate our approaches in real-world scenarios. 
\subsubsection{Quadrotor Inertial Dataset}
\begin{figure}[h!]
	\begin{center}
	\includegraphics[width=\columnwidth]{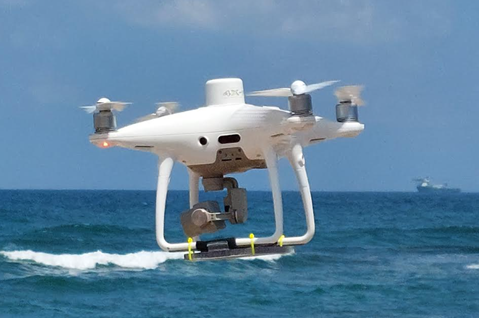}%
	\end{center}
	\caption{DJI Phantom 4 RTK equipped with Movella Dot IMUs during our recording campaign.}
	\label{Phantom4}
\end{figure}

We employ our recent dataset, published as part of our quadrotor dead reckoning research\cite{hurwitz2023quadrotor}. For the recordings, a DJI phantom 4 RTK drone equipped  with four Movella Dot IMUs, as shown in Figure \ref{Phantom4}, was used. The GT data was taken from the quadrotor build in inertial and GNSS RTK solution, while the Movella Dot IMUs recordings were the units under test. The Movella DOT sensor is small and lightweight sensor with an easy setup using Bluetooth connection. The Movella Dot is based on a BOSCH BNO055 MEMS IMU  consisting of three-axes accelerometer, gyroscope, and magnetometer.  For our analysis, only the accelerometer and gyroscope measurements were employed. The gyroscope specifications~\cite{Movella} state an in-run bias stability of $10deg/h$ and noise density of $0.007(deg/s/)\sqrt{Hz}$.  The accelerometer in-run bias stability is $0.03 mg$ and the noise density is $120 \mu g/\sqrt{Hz}$. The sampling rate was $120$Hz.
The recorded dataset contains straight-line and horizontal PTS trajectories. The former trajectories are used for comparison to the baseline INS solution. They include four trajectories with a total time of 66 seconds with an average quadrotor speed of 5.4 m/s. In addition, 27 PTS trajectories were recorded with a total duration of 16.1 minutes with an average quadrotor speed of 3.7 m/s. A typical PTS trajectory has an average length of 144m and takes 36s to complete. For each IMU, the training set consisted of 22 PTS trajectory recordings with a total time of 12.8 minutes, while the testing set had 5 PTS trajectories (not present in the train) with a total time of 3.3 minutes.
\subsubsection{Mobile Robot Inertial Dataset}
\begin{figure}[h!]
	\begin{center}
	\includegraphics[width=\columnwidth]{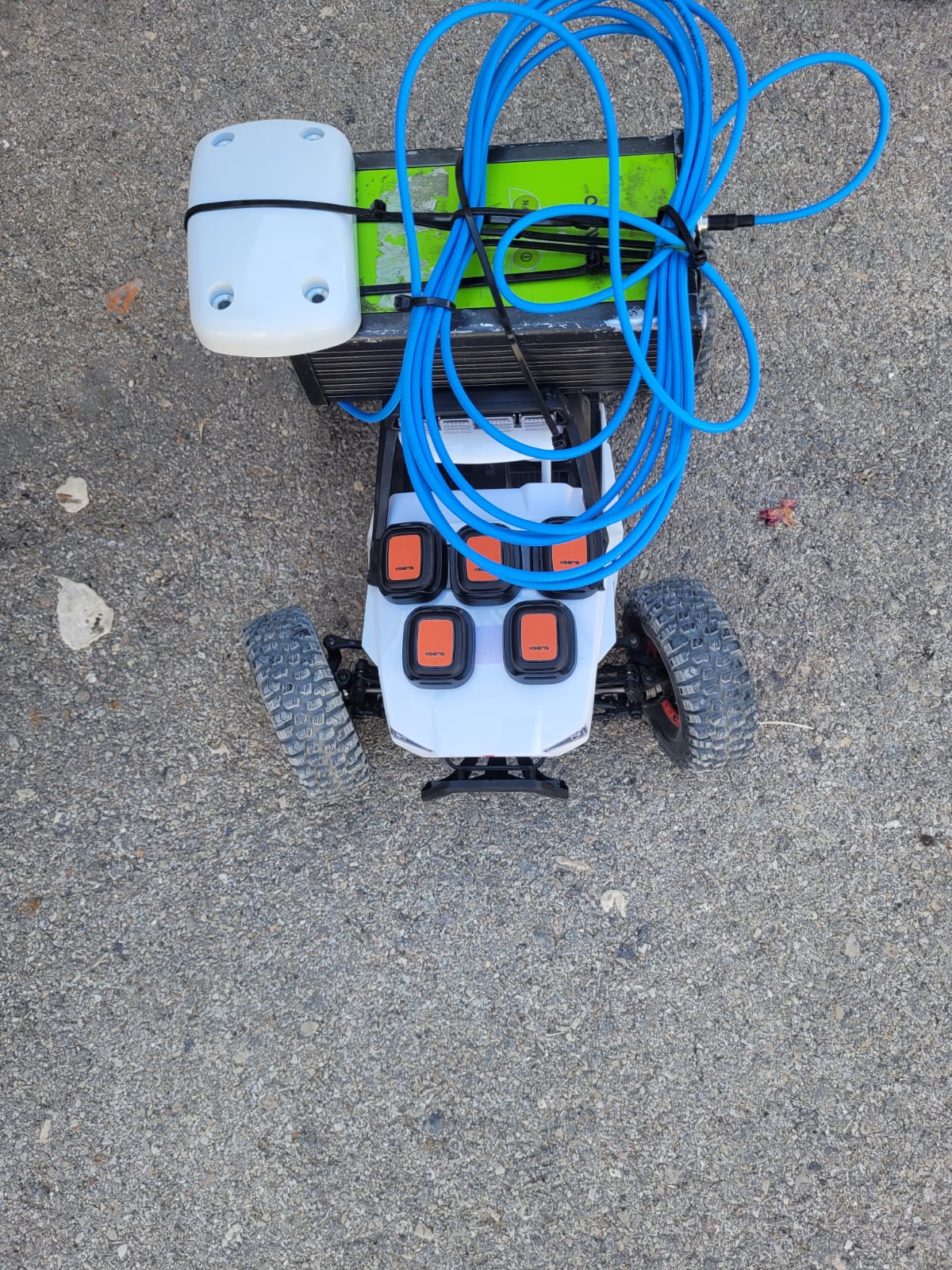}%
	\end{center}
	\caption{STORM Electric 4WD climbing car equipped with five Movella Dot  IMUs and a Javad GNSS RTK during our recording campaign.}
	\label{STORM}
\end{figure}

For the mobile robot recordings, we utilized a STORM electric 4WD climbing radio-controlled (RC) car with dimensions of 385x260x205 mm, a wheelbase of 253 mm, and a tire diameter of 110 mm. On top of the RC car, we mounted a Javad SIGMA-3N GNSS RTK sensor~\cite{Javad} to provide high-precision GT trajectories with a $10$cm accuracy while sampling in $10$Hz. Additionally, five Movella DOT IMUs were rigidly attached to the car's front. Our RC car experimental setup is presented in Figure~\ref{STORM}.
The car followed various predefined trajectories, including  14 PTS and three straight-line trajectories. The total recording time of the mobile robot inertial dataset for a single IMU is 67 minutes, including 5.5 minutes of straight-line trajectories.
The training set consists of nine PTS  recordings with a total time of 55 minutes and the testing set consists of five PTS  recordings with a total time of 6.4 minutes.
In a typical mobile robot PTS, the average length is 133m, and the average time taken to complete the trajectory is 223 seconds. We note that one out of the five IMUs experienced some malfunctions during the recording campaign. Therefore, only four IMUs were used in our evaluation and analysis.
To summarize, the datasets include 49 different PTS with 82.2 minutes of recording for each IMU. The train/test dataset duration (only PTS) is given in Table~\ref{Tbl:DatasetSplit}. 
\begin{table}[h!]
\begin{center}    
\caption{Duration of train and test datasets}
\label{Tbl:DatasetSplit}
\resizebox{0.4\textwidth}{!}{%
\begin{tabular}{|>{\color{black}}c|>{\color{black}}c|>{\color{black}}c|}
\hline
      & Quadrotor & Mobile Robot\\ \hline
PTS Train  [minutes] &      12.8    &         55.0           \\ \hline
PTS Test  [minutes]  &           3.3 &         6.4          \\ \hline
Straight line Test  [minutes]  &    1.1       &         5.5          \\ \hline
\end{tabular}%
}
\end{center}
\end{table}
\subsection{Performance Metrics}
\noindent
The proposed deep-learning assisted inertial dead recooking and fusion approaches are evaluated using two metrics: the position root mean square error (RMSE) and the position maximum error along the trajectory. The RMSE is defined by:
\begin{equation}
	\mathbf{RMSE}(\bold{p_i}, \hat{\bold{p}}_i) = \sqrt{\frac{\sum^N_{i=1}(||{\bold{p_i}-\hat{\bold{p}}_i}||)^2}{N}}
\end{equation}\label{eq:PRMSE}
where \textit{N} is the number of samples, $\bold{p_i}$ is the GT position vector, and $\hat{\bold{p}}_i$ is the estimated position vector. \\
In addition, to evaluate the regression network performance  (outputs the change in distance or altitude), we employ the distance RMSE  (DRMSE) defined as:
\begin{equation}\label{eq:DRMSE}
	\mathbf{DRMSE}({d_i}, \hat{{d}}_i) = \sqrt{\frac{\sum^N_{i=1}({{d_i}-\hat{{d}}_i})^2}{N}}
\end{equation}
where ${d_i}$ is the GT change in distance, and $\hat{{d}}_i$ is the estimated change in distance. 
\subsection{Experimental Results}
\subsubsection{PTS Trajectories for INS/GNSS Fusion}
Generally, when the platform maneuvers, the observability of the system improves, leading to accurate estimation\cite{bar2011tracking}. Using the test datasets of the quadrotor and mobile robot and the straight-line recordings, we examine the rate of improvement of PTS compared to straight-line trajectories. In both trajectory types, INS/GNSS fusion is applied using the error-state EKF \eqref{EKFIC}-\eqref{eq:CovarinaceUpd} with 1Hz GNSS position updates. As our goal in this part is to evaluate the trajectory types without the MQN approach, both training and testing datasets are employed for the evaluation.   Table~\ref{TrajShape} gives the position RMSE of the different trajectory types in the two datasets for all four IMUs. Notice that a slight improvement of PTS over straight-line trajectories is evident with $3.3\%$ improvement in the quadrotor dataset and $4.5\%$ in the mobile robot dataset. The standard deviation of the improvement is less than $0.3\%$ for both platforms. 
\begin{table}[h]
\caption{INS/GNSS fusion with 1Hz GNSS position updates.}
\begin{center}    
{%
\begin{tabular}{|c|c|c|}
\hline
  & \begin{tabular}[c]{@{}c@{}} Quadrotor\end{tabular} & Mobile Robot  \\ \hline
Straight line RSME {[}m{]}& 5.46 & 6.12  \\ \hline
PTS RMSE {[}m{]} & 5.26 & 5.85  \\ \hline
Improvement {[}\%{]}&3.3&4.5 \\ \hline
\end{tabular}%
\label{TrajShape}}
\end{center}
\end{table}
\subsubsection{Mini-QuadNet Evaluation} \label{MQN--Distance}
The test dataset of the quadrotor is used to evaluate the proposed MQN network compared to the QuadNet baseline using the DRMSE~\eqref{eq:DRMSE} metric. Table \ref{SmallQuadNet} shows an improvement of $20\%$ compared to the baseline. As a result, MQN not only requires $88\%$ fewer parameters and provides faster inference time but also improves distance regression performance.
\begin{table}[h]
\centering
\caption{Distance performance based on the DRMSE metric applied to the quadrotor testing data.}
\setlength{\tabcolsep}{3pt}
\begin{tabular}{|c|c|c|c|}
\hline
 & QuadNet (Baseline) & MQN (ours) & Improvement [\%]\\
\hline
DRMSE  [m]& 0.70& 0.56 & 20\\ \hline
Max error [m] & 1.42 & 1.3 & 8.5 \\ \hline
\end{tabular}
\label{SmallQuadNet}
\end{table}
\subsubsection{MQN/INS Fusion}
When the GNSS signals are not available, the platform operates in pure inertial navigation. For the quadrotor four straight line trajectories, pure inertial navigation result in a position RMSE of $226$m on an average trajectory length of $89$m. An RMSE error of $1,832$m was obtained for the mobile robot. This result is higher than the quadrotor RMSE because of longer trajectory duration (3.5 times longer) and the undamped motion of the robot on the ground. 
To mitigate this drift we examine two possibilities: 1) MQN: a dead reckoning approach using MQN as described in \eqref{eq:pos_x}-\eqref{eq:pos_z} and 2) MQN-EKF: a neural-inertial approach by updating the navigation filter with MQN solution using \eqref{eq:pos_mqn_fil}-\eqref{eq:HQuadNet}. \\
\noindent
The position RMSE, presented in Table~\ref{INS-QuadNet}, is obtained by combining the results from all four IMUs, each trained and tested separately. Both MQN and MQN-EKF greatly improved the INS baseline. In the quadrotor dataset, this improvement was between $94-95\%$. Additionally, our MQN-EKF approach improved by $12\%$ the MQN method. That is, the addition of the regression-based position as an update to the filter contributed to improving the positioning accuracy. \\
\noindent
In the mobile robot dataset, our approaches improved the baseline by $99\%$. However, the MQN-EKF obtained more or less the same performance as the MQN approach. This means that the MQN updates did not help in improving the position RMSE using the mobile robot. Further analysis shows that some of the IMUs experienced high fluctuations compared to others (and also to those mounted on the quadrotors). Our hypothesis is that this is due to the surface on which the experiment was conducted and a possible lack of rigid mounting of several IMUs. \\
\begin{table}[h]
\centering
\caption{RMSE errors for INS, MQN, and MQN-EKF with 1Hz MQN updates.}
\setlength{\tabcolsep}{3pt}
\begin{tabular}{|c|c|c|}
\hline
& 
Quadrotor & Mobile Robot  
 \\
\hline
INS (baseline) RMSE [m] & 226&  1,832 \\ \hline
MQN (ours)RMSE [m]&11.5 &13.0 \\ \hline
MQN-EKF (ours) RMSE [m] &10.1 & 13.2\\ \hline
Improvement relative to MQN [\%] & 12.0 & -1.4\\ \hline
\end{tabular}
\label{INS-QuadNet}
\end{table}
\subsubsection{MQN/INS/GNSS Fusion} \label{sec:Results and discuessions}
\begin{figure}[h!]
	\begin{center}
		\includegraphics[width=\columnwidth]{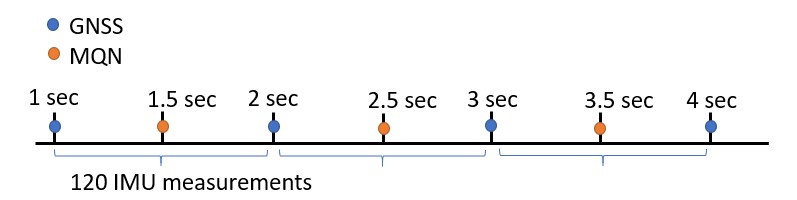}%
	\end{center}
	\caption{Position update timing for GNSS and MQN. }
	\label{Fig:Update}
\end{figure}

\begin{figure}[h!]
\begin{center}
    	\includegraphics[width=\columnwidth]{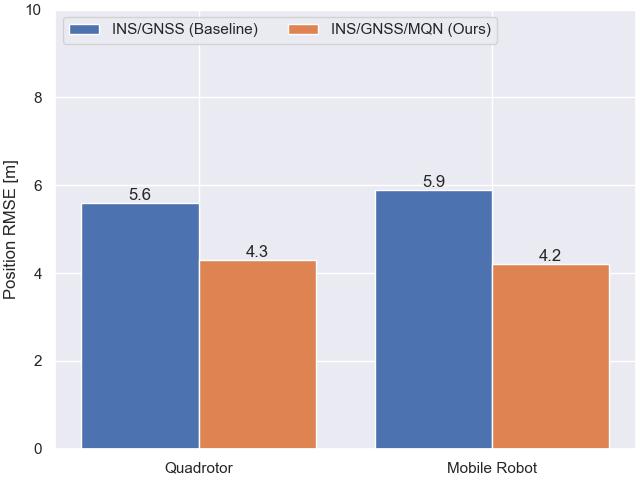}%
	\caption{Position RMSE without (baseline) and with (ours) MQN 1Hz updates.}
	\label{fig:posRMSEGNSS}
 \end{center}
\end{figure}

In this evaluation, our objective is to improve the commonly used INS/GNSS fusion by adding MQN updates. The idea is to use MQN updates in between GNSS updates. In our datasets both MQN and GNSS provides updates in 1Hz, thus we set the GNSS updates in each second and MQN updates, starting from 1.5s, in every second. The update timing procedure is illustrated in Figure~\ref{Fig:Update}.
\noindent
Figure~\ref{fig:posRMSEGNSS} provides the position RMSE of the INS/GNSS fusion (baseline) and INS/GNSS/MQN (ours) on the test dataset for all four IMUs. Notice that in Table~\ref{TrajShape} the INS/GNSS results were obtained for all the dataset (train and test) while here, for a fair comparison, only on the test dataset.\\
\noindent
In the quadrotor dataset, the addition of the MQN updates reduced the RMSE from 5.6 meters to 4.3, leading to a $21.5\%$ improvement, and in the mobile robot, the error decreased from 5.9 meters to 4.2, resulting in an improvement of $28.8\%$.
%
\section{Conclusion}\label{sec:con}
\noindent
This paper presented a lightweight network model, MQN, to regress the change in distance. Compared with the baseline model, it requires $88\%$ fewer parameters, provides faster inference time, and improves the positioning RMSE by $20\%$. This opens the use of low-cost hardware for real-time applications. Next, we derived an MQN-based position update to the navigation filter. It may be applied when GNSS is available as well as when it is not. To evaluate our approach, a 337 minutes dataset containing inertial measurements recorded while mounted on a quadrotor and on a mobile robot was employed. Due to their differing maneuvering capabilities, these two platforms were selected for our analysis in order to demonstrate the robustness of our approach. Furthermore, each platform was equipped with four IMUs, each of which was analyzed separately.\\
\noindent
We first showed that in GNSS/INS fusion, PTS can help in improving positioning accuracy. When GNSS is not available, the combination of PTS and our MQN dead-reckoning approach improves by more than $94\%$ for the quadrotor and mobile robot, the model-based pure inertial navigation approach. The reason for the high improvement lies in the fact that PTS enriches the inertial measurements with angular velocities and accelerations. This in turn, enables our MQN network to accurately estimate the change in distance of the platform. To further improve performance in GNSS-denied environments, we derived an MQN-based position update for the navigation filter. For the quadrotor, this update achieved a further improvement of $12\%$, while for the mobile robot, no improvement was made. This is attributed to the surface conditions and to the fact that several IMUs were not mounted rigidly enough on the platform. When GNSS position updates are available, we demonstrated that combined with MQN updates, the position accuracy is improved by $21.5\%$ for mobile robots and $28.8\%$ for the quadrotors.
In conclusion, combining PTS and our dead-reckoning MQN approach, both mobile robots and quadrotors are able to improve positioning accuracy in situations of GNSS outages and GNSS availability. 

%



\section*{Acknowledgment}
\noindent
This research was partly supported by the Ministry of Innovation, Science, and Technology under grant number 1001575651.
\ifCLASSOPTIONcaptionsoff
\newpage
\fi



%
{\bibliographystyle{IEEEtran}		\bibliography{WorkOnVersion.bib}}

\end{document}